\documentclass[10pt,a4paper,oneside]{article}

\usepackage{listings}
\usepackage{color}

\definecolor{codegreen}{rgb}{0,0.6,0}
\definecolor{codegray}{rgb}{0.5,0.5,0.5}
\definecolor{codepurple}{rgb}{0.58,0,0.82}
\definecolor{backcolour}{rgb}{0.95,0.95,0.92}
 
\lstdefinestyle{mystyle}{
    backgroundcolor=\color{backcolour},   
    commentstyle=\color{codegreen},
    keywordstyle=\color{magenta},
    numberstyle=\tiny\color{codegray},
    stringstyle=\color{codepurple},
    basicstyle=\footnotesize,
    breakatwhitespace=false,         
    breaklines=true,                 
    captionpos=b,                    
    keepspaces=true,                 
    numbers=left,                    
    numbersep=5pt,                  
    showspaces=false,                
    showstringspaces=false,
    showtabs=false,                  
    tabsize=2
}
 
\usepackage[utf8]{inputenc}
\usepackage{verbatim}
\usepackage[T1]{fontenc}
\usepackage{amsmath}
\usepackage{amsfonts}
\usepackage{amssymb}
\usepackage{graphicx}
\usepackage{float} 
\usepackage[left=2cm,right=2cm,top=2cm,bottom=2cm]{geometry}
\usepackage{multicol}
\usepackage{amsthm}
\usepackage{textcomp}
\usepackage{hyperref}
\usepackage{babel}
\usepackage{listings}
\usepackage{csquotes}
\lstdefinelanguage{IAChat}{
    basicstyle=\ttfamily,
    keywordstyle=\color{blue}\bfseries,
    commentstyle=\color{gray}\itshape,
    morecomment=[l][\color{gray}\itshape]{[MODEL] :},
    morecomment=[l][\color{orange}]{[PROMPT] :},
    frame=single,
    numbers=left,
    xleftmargin=10pt,
}

\author{
Thibault, Chanus \\
 \texttt{thibault.chanus@ens-rennes.fr} \\
 \textit{Internship ENS Rennes} \\
   \and
Michael, Aubertin \\
 \texttt{maubertin@fr.scc.com}\\
 \textit{Staff Advisor SCC France} \\
}

\title{Exploring the Application of LLM in Infrastructure as Code}

\begin{document}
\maketitle

\vspace{0.8cm}

\begin{figure}[h!]
\centering
\includegraphics[scale=0.3]{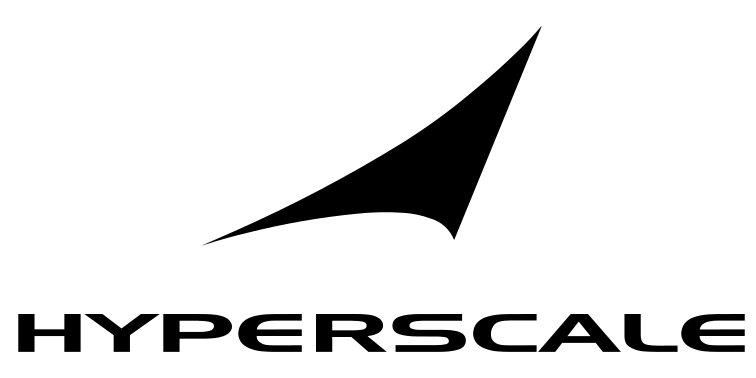}
\end{figure}

\vspace{0.8cm}

\begin{abstract}
Cloud computing and the evolution of management methodologies such as Lean Management or Agile entail a profound transformation in both system construction and maintenance approaches. These practices are encompassed within the term "DevOps." This descriptive approach to an information system or application, alongside the configuration of its constituent components, has necessitated the development of descriptive languages paired with specialized engines for automating systems administration tasks. Among these, the tandem of Ansible (engine) and YAML (descriptive language) stands out as the two most prevalent tools in the market, facing notable competition mainly from Terraform\texttrademark. The current document presents an inquiry into a solution for generating and managing Ansible YAML roles and playbooks, utilizing Generative LLMs (Language Models) to translate human descriptions into code. Our efforts are focused on identifying plausible directions and outlining the potential industrial applications. 
\end{abstract}

\vspace{0.8cm}

\begin{minipage}{0.95\textwidth}
\smallskip
\textbf{Note :} For this experimentation, we have consciously elected not to leverage Ansible Lightspeed. The rationale behind this choice stems from the unavailability of any documented sources pertaining to the IBM\texttrademark Watson\texttrademark model upon which Ansible Lightspeed is predicated. You may access the pertinent information~\cite{redhat} regarding this cutting-edge technology directly on the website of our distinguished partner, RedHat\texttrademark.
\end{minipage}

\vspace{1.2cm}

\begin{multicols}{2}
\section{Introduction}
With the surge of cloud computing and the increasingly widespread adoption of modern management methodologies such as Lean Management and Agile, the information technology landscape has undergone a radical metamorphosis. These developments have led to the emergence of a new approach to development and operations, commonly referred to as DevOps. The core objective of this approach is to amalgamate the realms of development and operations, thereby augmenting the efficiency and quality of information systems and applications.
\\
In this context, the precise and automated description of systems and applications has become imperative. It serves not only to delineate the configuration of their constituent components but also to automate associated administrative tasks. To address this demand, descriptive languages accompanied by automated processing engines have been developed. Among these instruments, the Ansible (in the role of engine) and YAML (in the capacity of descriptive language) tandem distinguish themselves as one of the preeminent solutions in the market, with Terraform as its chief competitor.
\\
Nonetheless, the creation and management of YAML files for Ansible can pose complexity, particularly for non-technical users. The necessity for a more intuitive and innate interface for the generation of these files has, consequently, emerged. It is within this context that the employment of Large-Scale Language Models (LLMs) has been considered as a potential solution. LLMs are profound machine learning models, having undergone training on extensive corpora of textual data, thus endowed with the capability of text generation. They have, withal, been efficaciously harnessed across diverse applications, inclusive of code generation.
\\
The primary aim of this study resides in the exploration of Large Language Models (LLMs) capacity to confront this challenge. Exploiting the intrinsic potential of LLMs to decipher natural language and generate code, our endeavor is to fashion a solution that may facilitate and mechanize the crafting of YAML files for Ansible, thereby endowing the process with enhanced accessibility and efficiency.
\\
The potential utilization of LLM to generate YAML-based Ansible files from human textual descriptions ushers in a new epoch of automation and efficacy within the DevOps domain. In the subsequent sections, we shall expound upon our methodology, proffer our preliminary findings, and engage in a discourse concerning the prospective implications of our revelations for the industry.
\end{multicols}

\vspace{0.8cm}

\begin{multicols}{2}
\section{Technical Overview}
Within this chapter, we present a brief overview of the functioning of the various key elements utilized by the LLM models under scrutiny in our experiments.

\end{multicols}
\subsection{The Transformer Architectur}

In recent years, and more notably, in recent months, the Transformer architecture has witnessed a substantial evolution. While it retains certain incompleteness and presents challenges in its accessibility, the definitive reference for this architecture stems from the research laboratories of Google ~\cite{vaswani2023attention}. Presented here are some pertinent reminders to enhance comprehension of our experimental endeavors.

\begin{multicols}{2}

\subsubsection{History of Transformer}
\begin{figure}[H]
\centering
\includegraphics[width=0.5\textwidth]{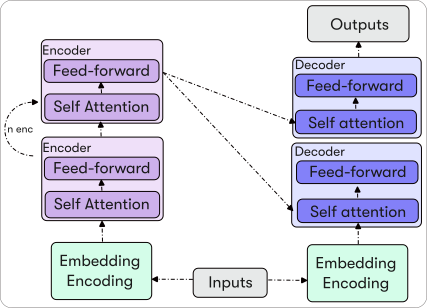}
\caption{Conceptual Diagram of the Transformer Architecture}
\label{fig:schemaprinctransfo}
\end{figure}

Historically, the Transformer architecture is comprised of an encoding block and a decoding block (Refer to \ref{fig:schemaprinctransfo}).

In its early stages, this model emerged as the most suitable choice for its primary application, which was text translation. Since then, research has progressed, and applications have diversified, now prioritizing specialized functionalities, necessitating either encoding, as in information retrieval, or decoding, as in text generation. Nevertheless, the need for models capable of mobilizing both an encoder and a decoder persists. Our specific use case pertains exclusively to text generation, which means that our models exclusively leverage the Decoder component.

The illustration below ~\cite{yang2023harnessing}, provides a brief glimpse, dating back to early 2023, of the proliferation of implementations of the Transformer architecture. Our experiments and analyses lead us to observe that in just 8 months, major models have already rendered their own obsolescence conspicuously evident.
\end{multicols}

\begin{figure}[H]
\centering
\includegraphics[scale=0.5,width=1\textwidth]{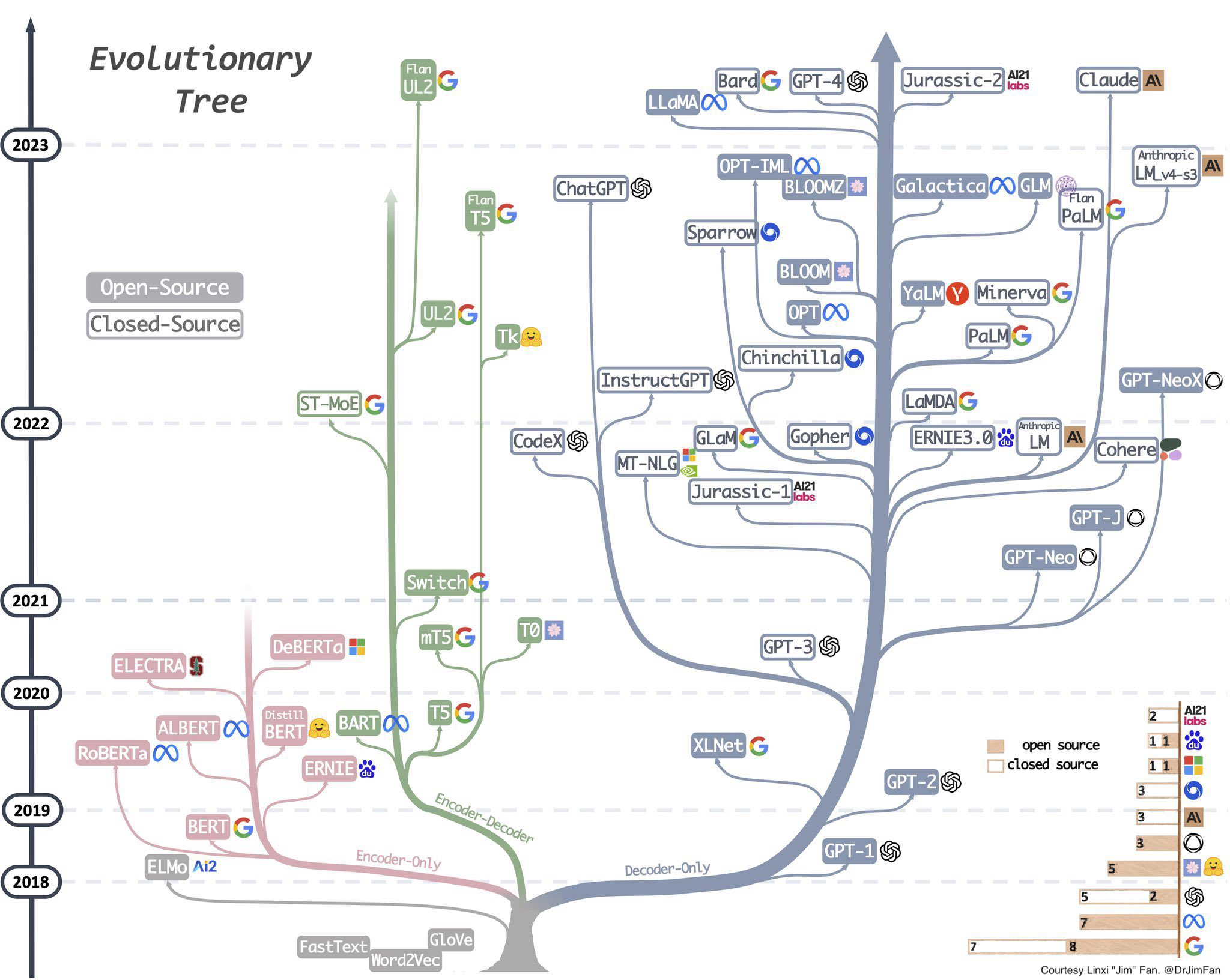}
\caption{Evolution tree of models over the last 5 years}
\label{fig:evol_transfo}
\end{figure}

\begin{multicols}{2}
\subsubsection{Tokens and the Tokenizer}
In practical terms, a tokenizer's mission is to translate a sentence into a sequence of numbers. To accomplish this task, it generates "tokens," which are numerical representations of groups of symbols. The tokenizer's true efficiency lies in its ability to identify the most frequently used symbol groups and organize them optimally. Consequently, it assigns a unique number to each symbol group, with their sizes being variable. The primary challenge of a quality tokenizer lies in its capacity to determine the appropriate grouping size for sets of symbols (words).
\\
Each model possesses a fixed token-processing capacity. A symbol group size that is too small will disrupt the model's performance, as it will consume an excessive number of input tokens relative to the input text size processed by the tokenizer. Conversely, a symbol group size that is too large will diminish the model's relevance, as the abundance of distinct tokens will hinder the model's ability to comprehend and learn the meaning of each token. The model acquires its performance through its presumed understanding and learning during the training phase.
\\
Additionally, the tokenizer enables us to derive meta-tokens, which serve the purpose of signaling its operation. For example, the end-of-sentence signal (EndOfSentenceToken). Consequently, the tokenizer is an essential component that should be viewed as intrinsic and inseparable from a model.

\subsubsection{The context}
Before delving deeper into the intricacies of our approach, it is crucial to grasp a fundamental concept of LLMs: context.

In the context of LLMs, "context" refers to the size of the token window that the model takes into account when generating the next token. In simpler terms, it signifies the amount of immediately preceding information that the model uses to predict the subsequent text.

It is imperative to understand that the primary objective of an LLM is to predict the next token based on preceding tokens. However, due to computational and structural limitations, the model can only consider a limited number of tokens at once. This constraint is what we refer to as the "context window." Any information beyond this window does not influence the model's prediction for the next token.

\end{multicols}

\subsubsection{The Self Attention}
The self-attention mechanism, in the field of artificial intelligence, is designed to facilitate a model in effectively handling the relationships between different segments of a sequence, such as words in a sentence. In practice, self-attention empowers the model to ascertain the extent to which each element in a sequence is interconnected with the others, assigning pertinent weights to these connections.

The core concept behind self-attention is to generate "attention representations" for each element in the sequence. These representations reflect the relative importance of other elements concerning the element in question. To achieve this, self-attention employs three distinct linear projections, referred to as key, query, and value transformations. These transformations convert each element into three vectors, which are then utilized to compute attention scores between all pairs of elements.

Subsequently, these attention scores are normalized using functions such as the softmax function, resulting in normalized attention weights. These weights indicate the extent to which each element should prioritize other elements in the sequence. Higher weights are assigned to elements with stronger connections to the current element under examination.

Once attention weights are obtained, they are used to combine the value representations of other elements. This weighted combination of values yields a final representation that captures the contextual relationships between sequence elements.

In summary, self-attention is a technique that enables a model to effectively address long-range dependencies among sequence elements by computing attention weights for each pair of elements. These weights are then employed to generate a new contextual representation for each element, thereby enhancing the model's capacity to comprehend intricate relationships within the sequence.

Within the Transformer architecture, we thus obtain a representation, as illustrated here:

\begin{figure}[H]
\centering
\includegraphics[scale=0.4,width=0.8\textwidth]{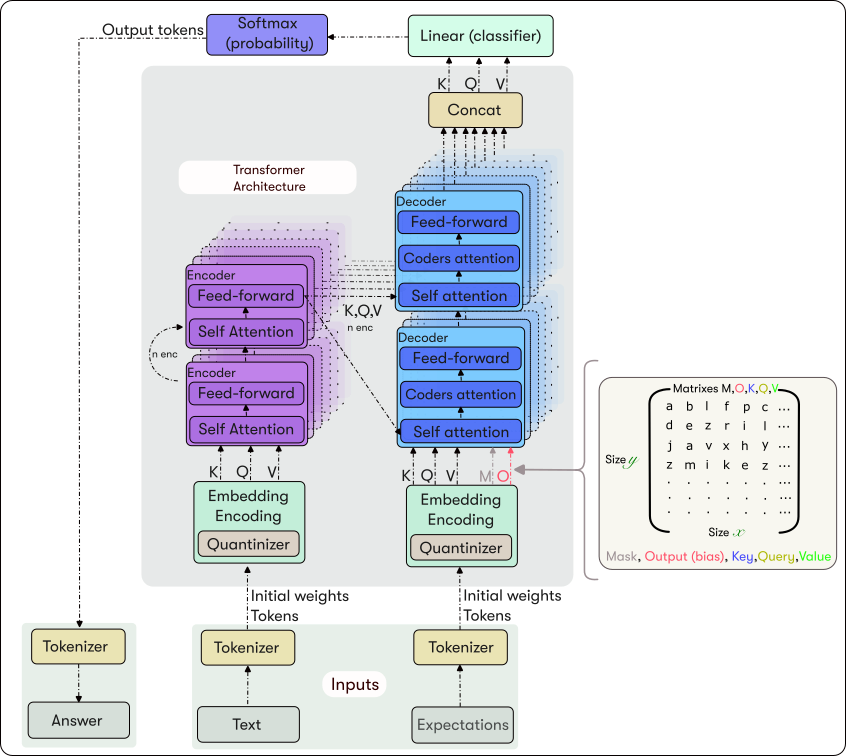}
\caption{Transformer Architecture and Usage Model}
\label{fig:det_transfo}
\end{figure}

With its functioning mechanism being:
\begin{normalsize}
$ MultiHead(Q, K, V ) = Concat(head_{1}, ..., head_{n}).W^{O} \\
   With: head_{i} = Attention(QW^{Q}_{i},KW^{K}_{i},VW^{V}_{i}) \\
And\ weight\ matrixes\ (Q,K,V,O)\ as:\ \\
W^{Q}_{i} \in R^{dmodel \times d_{k}},\ W^{K}_{i} \in \mathbb{R}^{dmodel \times d_{k}},\ W^{V}_{i} \in \mathbb{R}^{dmodel \times d_{v}},\ W^{O} \in \mathbb{R}^{hd_{v} \times d_{model}} \\
Et\ i \in [{0,n^{Heads}}] \\
$
\end{normalsize}

It should be noted that it is also common to seek to optimize the attention model by reducing distance computations. In the models we have experimented with, a technique known as Shuffle is used. It relies on introducing randomness into the calculation results to more effectively evaluate distances by comparing weight scores, irrespective of the sequence order ~\cite{huang2021shuffle}.

\vspace{0.8cm}

\begin{multicols}{2}
\subsubsection{Feed-forwarder}

When applied to the Transformer architecture, the concept of feed-forwarder is a pivotal element in the functioning of this artificial intelligence model. The feed-forward mechanism takes inspiration from the operation of neural memory within the human brain ~\cite{sukhbaatar2019augmenting}. It empowers the model to carry out real-time data processing operations by employing layers of straightforward computations. To delve into the intricacies of the feed-forward mechanism within the Transformer architecture, it is imperative to take into account the following factors.

Le concept général de feed-forward appliqué à l'architecture Transformer constitue un élément clé dans le fonctionnement de ce modèle d'intelligence artificielle. Le feed-forward s'inspire du fonctionnement de la mémoire neuronale dans le cerveau humain ~\cite{sukhbaatar2019augmenting}. Il permet au modèle de réaliser des opérations de traitement instantanées sur les données en utilisant des couches de calcul simples. Pour comprendre en détail le mécanisme du feed-forward au sein de l'architecture Transformer, il est essentiel de prendre en considération les éléments suivants.

In contrast to more intricate processing mechanisms, the feed-forward mechanism centers around consecutive linear transformations of data, rendering it an efficient process for a wide range of tasks.

The feed-forward process operates by employing the self-attention mechanism on data with values ~\cite{geva2021transformer}. This results in an action on the key (K), query (Q), and value (V) matrices derived from the previous self-attention phase. The Q and K matrices are utilized to compute attention scores, while the V matrix holds the information for correlation. However, unlike the self-attention stage, where all interrelationships among elements are taken into account, the feed-forward has the option to selectively process information, thereby optimizing efficiency.

The feed-forward concept can be realized in various manners within the Transformer architecture. Different variations of activation functions, layer combinations, and optimization techniques can be harnessed to execute processing operations, as depicted in the following figure.

\end{multicols}
\begin{figure}[H]
\centering
\includegraphics[scale=0.8]{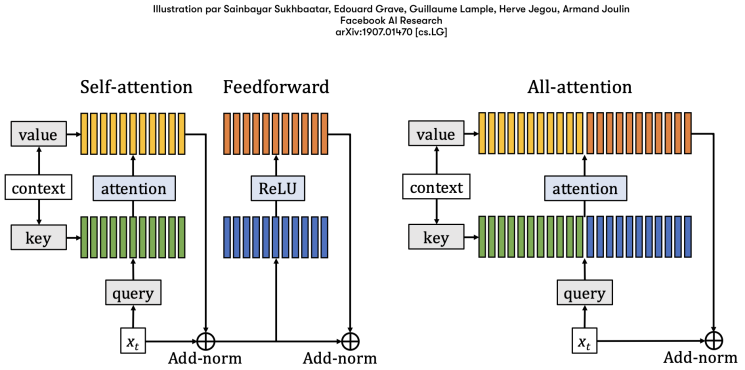}
\caption{Visualization of the functioning of the Feed Forwarder (Neural Memory)}
\label{fig:det_feed}
\end{figure}

\begin{multicols}{2}

As an example, although the original model of the Transformer architecture was conceived at Google (employing activations such as ReLU and the mechanisms of Multi-Head Self-Attention), implementation alternatives like GLU (Gated Linear Unit) ~\cite{shazeer2020glu} can also be embraced to improve both model performance and convergence.

The LLMA (Large Language Model Architecture) approach, although originating from Meta, presents an intriguing adaptation of the feed-forward mechanism. This adaptation employs a variant of GLU, as endorsed by Google. The objective of this approach is to enhance the model's performance by harnessing the benefits of GLU in capturing contextual dependencies. It is noteworthy that, despite Google's recommendations emphasizing activations such as GEGLU (Gated Exponential Linear Unit) or ReGLU (Rectified Gated Linear Unit), the LLMA code demonstrates the utilization of the "silu" activation function ~\cite{Sigmoid}, also referred to as the Sigmoid Linear Unit, to fulfill a similar role.

In essence, the application of the feed-forward concept to the Transformer architecture allows for the emulation of neural memory, carrying out immediate data processing operations through consecutive linear transformations. It utilizes the Q, K, and V matrices from the self-attention phase and can be implemented in diverse ways to enhance model performance, as exemplified by the LLMA approach with its GLU variant and the incorporation of the "silu" function.

The function of such a feed-forward is thus in the following form:

$$
y_{t} = \sum_{c \in C_{t}} a_{tc}(V_{c} + p(t,c)) \ et \  a_{tc} = \frac{exp (S_{tc}/\sqrt{d_{h}})}{\sum_{i \in C_{t}}exp (S_{ti}/\sqrt{d_{h}})}
$$

\subsection{Quantization: Mastering the Optimization of Massive Language Models (LLMs)}

In the ever-evolving domain of Large-Scale Language Models (LLMs), quantization~\cite{quantzmode} emerges as a sophisticated and indispensable optimization strategy. Historically, while the float32 format has been the favored standard for LLM model weights, it has become evident, through relentless research and development, that other formats can deliver superior performance. The transition to float16 marked a significant milestone by offering a more compact structure. However, the true breakthrough materialized with the adoption of formats such as int8 or int4. We shall overcome this complexity, not because it is easy, but because it is hard! LLM models function in a simplified manner by harnessing numerical variations within matrices. For a considerable period, research presumed that the precision in measuring these variations was of utmost importance. Recent advancements in knowledge demonstrate that such precision does not bear crucial significance. Therefore, akin to the well-established quantization process employed in signal digitization methods, certain iterations of the most contemporary models incorporate a quantization step~\cite{frantar2023gptq}. This diminishes the necessity for high precision in every numeric value. This approach of reduced precision facilitates the utilization of integer variables (int), which, from a computational perspective, proves more advantageous than employing floating-point variables, renowned for their resource-intensive demands. Such meticulously calibrated models necessitate less storage space and seamlessly align with the most performant assembly instructions across all known architectures.

These (int) data formats, while significantly reducing memory usage, greatly enhance computation speed. Opting for integers, as opposed to floating-point values, represents a strategic approach that capitalizes on optimized instructions for calculations, while minimizing the space occupied by data. In the pursuit of a rigorous optimization of computational resources, it is imperative to recognize the importance and relevance of the quantization methodologies that currently dominate the industry:

- Dynamic Quantization: In this mode, weights undergo quantization to integers prior to inference. However, since the neural network lacks knowledge of the scales and zero points of the output or activation tensors, these tensors remain in floating-point form. The advantage lies in the fact that no data calibration is necessary before inference.

- Static Quantization: Here, the scales and zero points of all activation tensors are precomputed, thus eliminating the overhead of dynamic computation. The process involves running the neural network with representative unlabeled data, collecting distribution statistics for activation layers, and then calculating the scales and zero points. This mode provides the fastest inference performance but requires representative unlabeled data.

In order to accurately comprehend the impact of quantization on model size, we provide an illustrative table below. It is essential to note that the figures mentioned are approximate. They solely reflect the model weight post-transformation, without taking into account the weight of transformation parameters. The latter can be finely tuned to optimize the overall model performance by adjusting the quantization window.

\begin{figure}[H]
\centering
\includegraphics[scale=0.3,width=0.5\textwidth]{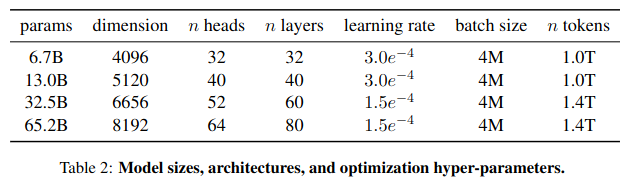}
\caption{Model sizes,architectures,optimization hyper-parameters}
\label{fig:det_llama1}
\end{figure}
The principle of quantization can be applied during the training phase and during the tuning phase. However, it is generally acknowledged that quantization during the training phase yields better results. Nevertheless, the computational power at our disposal for our experiments has only allowed us to perform quantization during the tuning phase.

In our case, quantization, embedded within the transformer, is applied to all weight matrices of the attention model, as well as to the output embedding layer of the tokenizer and the so-called Feedforward neural layer.

Le principe de quantification peut être appliqué en phase d'entrainement et en phase de tuning. Il est toutefois généralement admis qu'une quantification en phase d'entrainement donne de meilleurs résultats. Toutefois, la puissance de calcul à notre disposition pour nos expérimentations ne nous a permis que la quantification en phase de tuning.

Dans notre cas, la quantification, embarquée dans le transformer, s'applique à toutes les matrices de poids du modèle d'attention ainsi qu'à la couche d'embarquement en sortie de tokenizer et à la couche neuronale dite Feedforward. The quantization equation used for 4 and 8 bits is as follows:

\begin{align*}
    \scriptstyle quantize(W_{fp},n) = round(((W_{fp} - min(W_{fp})) \times \frac{2^n - 1}{max(W_{fp}) - min(W_{fp})}) \\
    \scriptstyle With \ W_{fp} \ the \ K,Q,V \ matrixes \ in \ float32 \\
    \scriptstyle And \ let \ n \ be \ the \ quantization\  factor.
\end{align*}

For self-attention models that utilize the matrices Q, K, V, as follows:
\begin{align*}
    Attention(Q,K,V) = softmax(\frac{QK^T}{(\tau) \sqrt{d_{k}}})V
\end{align*}

Note: The significance of $\tau$ is relevant only when quantization is applied to a single bit. We have not conducted any experiments involving such a level of quantization.

The latest available publications indicate that an 8-bit integer quantization approach offers model performance extremely close to that of the original 32-bit floating-point model ~\cite{dettmers2022llmint8}. In addition to its efficiency and processing simplicity, this 8-bit approach reduces the model size by a factor of 4. Therefore, the potential of this approach is considerable.

For more details, please refer to these publications ~\cite{quantzmode} and ~cite{quantSOTABERT}.


\vspace{0.5cm}
\section{Difficulties associated with automatically generating Ansible YAML files using LLMs.}
The automatic generation of Ansible YAML files from human textual descriptions using Large Language Models (LLMs) presents a unique set of challenges. These challenges, which encompass the creation of the file itself and the quality of the generated code, are fundamentally important to grasp in order to develop a workable solution. In this section, we will elaborate on the primary difficulties associated with this task.

\subsection{File generation by the LLM.}

The first challenge lies in the creation of the YAML file itself by the LLM. This entails not only finding the right wording of the query to the LLM (prompt) to achieve the best results but also selecting a suitable model. The chosen model must be able to operate within the specified constraints and generate the desired content efficiently. The "prompt engineering" challenge is especially critical as the initial prompt largely determines the quality and relevance of the generated text.

\subsection{Context capacity limitations}

LLMs based on the Transformer architecture, while formidable, possess inherent limitations in their contextual abilities. As an informative note, a well-crafted prompt can swiftly extend to encompass between 256 and 512 tokens and expands rapidly with the addition of examples or documentation. Given that expected responses may be extensive and demand numerous tokens, this significantly curtails the size of the initial prompt. This constraint substantially affects the quality and precision of the generated YAML file.

\subsection{What would be a "good" definition?}

A YAML file is inherently descriptive, often containing more information relative to its size than other types of documents. Consequently, defining what constitutes a "good" description for a YAML file is a challenge. An excessively detailed description renders the use of an LLM redundant, as converting such a description into a YAML file would be relatively straightforward. Conversely, an excessively concise description would not provide sufficient information to the LLM to generate an appropriate file. Optimizing the description of a file for YAML rendering is imperative. By constructing a diverse array of files and their descriptions, we could fine-tune the model for enhanced YAML generation. For businesses, this advancement translates to heightened reliability and improved operational efficiency.

\subsection{Handling of the generated data.}

Upon generating the file, its processing proves to be intricate. Essentially, what is produced is a lengthy string of characters. For a human, discerning the commencement and conclusion of the code is straightforward, yet for a model, this may prove more intricate. Certain models exhibit a tendency to encompass the code within two "```" sequences, thereby simplifying its identification. Nonetheless, this is not always the scenario, rendering the demarcation of the generated code more demanding. This circumstance assumes added significance, particularly since certain models are inclined to generate multiple code segments interspersed with explanations or comments, thus intensifying the complexity of automated processing.

\subsection{Quality of generated code}

The last, but certainly not the least, of the challenges pertains to the quality of the generated code. In our experiments, it has often been observed that the code may outwardly resemble YAML but lack internal consistency. Through multiple iterations and improvements, the code can gain greater coherence. However, subtle errors, such as a reference to a nonexistent module or incorrectly provided parameters, are not uncommon. In our Ansible use case, parameter errors arise due to the utilization of Ansible module libraries, each of which possesses and implies its unique grammar. Consequently, the YAML code generated must conform to these grammatical rules and conventions. Currently, the detection of these errors demands extensive YAML expertise and introduces complexity into the code generation process. With the ongoing enhancement of Large Language Models, especially in terms of context management and precision, we anticipate notable improvements in the quality of generated code, particularly in the context of YAML with Ansible. For enterprises, this advancement will result in more robust automation, ensuring high-quality code and reduced errors. Thus, it appears imperative to continue exploring the possibilities afforded by Large Language Models.

\vspace{0.5cm}
\section{vLLM}
\subsection{Introduction to Model}

vLLM stands as a technical solution crafted to facilitate interactions with Large-Scale Language Models (LLM). Its design is specifically tailored for seamless integration with Python, thus presenting an intuitive interface for developers.

The vLLM project is hosted on GitHub and is accompanied by comprehensive documentation accessible on ReadTheDocs~\cite{VLLM}. This documentation spans various topics, encompassing installation, rapid initiation, supported models, and also furnishes illustrative examples to assist users in their initial steps.

\subsection{Observations}

When assessing tools designed to facilitate work with Large Language Models (LLMs), vLLM emerges as an interesting option, notably due to its innate compatibility with Python. However, each solution harbors its distinctive features, and in vLLM's case, a noteworthy limitation surfaces in its inability to handle quantization. This characteristic confines its utilization to the LlaMa-7B model, given the specific hardware constraints of the project. It is also worth acknowledging that our available infrastructural resources for experimentation are ill-suited to accommodate a server version of vLLM, thus introducing an additional layer of limitation.

\subsection{Lesson}

The exploration of vLLM has highlight the critical significance of choosing appropriate tools in the field of artificial intelligence. Every tool, no matter how capable, carries its unique characteristics and constraints. vLLM, despite its robustness and ease of integration, has exhibited certain limitations that have impeded our ability to achieve satisfactory results for the project at hand. Thus, it is imperative, during the project's conception, to ensure that the selected tools align perfectly with the stipulated requirements and objectives.

\subsection{Conclusion}

vLLM seems to be a top-tier technical solution for engaging with Large-Scale Language Models, especially catering to Python enthusiasts. Its intuitive design, paired with comprehensive documentation, positions it as a prime candidate for numerous artificial intelligence projects. However, every tool, regardless of its sophistication, presents its unique set of challenges. Within the scope of our project, despite vLLM's initial promises, hardware constraints and model limitations have steered us away from adopting it as the primary solution.

\vspace{0.5cm}
\section{Llama cpp}
\subsection{Introduction to Model}
Llama CPP represents a C++ implementation of Llama, purpose-built for the execution of quantized LLM models (https://github.com/ggerganov/llama.cpp). This library presents a resilient substitute to prevalent Python implementations, yielding substantial advantages concerning performance and seamless integration. Although the integration with Python may prove more intricate, the gains in execution velocity and operational efficiency decidedly surpass this challenge.

\subsection{Observations}
The integration of Llama CPP into our Large-Scale Language Model (LLM) testing framework has unveiled several significant findings. To begin with, despite the initial challenges in usage, this approach has showcased its superiority in terms of performance. In fact, it was practically impossible to operate any model without employing some form of quantization. This observation is consistent with the information presented in the Llama CPP GitHub repository, emphasizing the importance of quantization in optimizing performance and resource utilization.

Moreover, a detailed and in-depth analysis has highlight a not so considerable difference in terms of relevance between the Int4 and Int8 quantified versions. This implies that, in certain applications, the shift from Int8 to Int4 can be carried out without jeopardizing the quality of results. Nonetheless, noticeable distinctions have arisen between the quantifications of model 7b and 13b, underscoring the imperative nature of selecting the right quantization approach to fully exploit the potential of a 13b model and enhance responses precision.

\subsection{Lesson}
The implementation of Llama CPP has yielded several valuable lessons. Quantization, despite potential losses stemming from simplifications, has emerged as a critical necessity for the functionality of LLMs in resource-limited scenarios. This underscores the imperative need for suitable tools and methodologies to govern and enhance LLM models, notably within settings characterized by constrained hardware resources.

The experience with Llama CPP has underscored the importance of documentation and training. While the library bestows notable performance advantages, its learning curve can prove steep for those unversed in the subtleties of LLM quantization and optimization. Hence, it is imperative to provide comprehensive documentation and practical examples to facilitate the library's adoption and integration into professional projects.

\subsection{Conclusion}

In conclusion, Llama CPP stands out as a robust technical solution for the implementation and optimization of LLMs. Its adoption requires a comprehensive understanding of quantization and its implications, but the advantages in terms of performance and efficiency make it a wise choice for companies seeking to fully leverage the potential of LLMs.

\vspace{0.5cm}
\section{LLaMA 7b}
\subsection{Introduction to Model}
The LLaMA 7B model from Meta AI ~\cite{touvron2023llama} distinguishes itself through its limited parameter count, a remarkable technical achievement in this field. It represents a notable advancement within the realm of Large Language Models (LLMs). This assembly of models, ranging from 7B to 70B parameters, has been trained on trillions of tokens. What sets LLaMA apart is its exclusive utilization of publicly available datasets. This open approach stands in contrast to other popular models, which often rely on proprietary or inaccessible data sources.

In our current professional paradigm, marked by the crucial significance of agility and fast execution, LLaMA 7B emerges as an optimal solution tailored for enterprises. Despite its comparatively reduced scale in contrast to other models such as GPT-3, this compactness, in reality, constitutes a major asset. LLaMA 7B, boasting its streamlined structure, facilitates faster execution and demands fewer resources, rendering it exceptionally suitable for environments where operational efficiency reigns supreme. Moreover, its training on open data bolsters its transparency, a criterion of increasingly high value.

\subsection{Observations}
Within the vast ecosystem of language models, LLaMA 7B stands out due to its compact size. Its reduced dimensions make it particularly attractive for deployments on resource-constrained machines, providing an unparalleled level of operational flexibility. This feature allows businesses to leverage artificial intelligence without the need for hefty or expensive infrastructures.

However LLaMA 7B's compact size comes with its own set of challenges. While it offers advantages in terms of operational efficiency, it poses limitations in processing capacity. When faced with tasks requiring profound comprehension or the generation of lengthy and nuanced responses, the model quickly reaches its limits. Moreover, it tends to be restricted by context size, making it less suitable for tasks demanding detailed analysis of extensive data volumes. Consequently, when producing more complex content, it may occasionally reiterate certain information, potentially compromising output clarity and precision.

A revelant sample of YAML code generation carried out by LLaMA 7B can be found in the appendices.

\subsection{Lesson}
When we approach the issue of generating Ansible files in YAML format, it is imperative to take into account the specifics and technical requirements associated with this task. Ansible, being a leading IT automation tool, relies on YAML files that must be meticulously structured and free from errors to ensure flawless execution.

In this context, our investigation of the LLaMA 7B model has unveiled certain significant nuances. Although this model possesses an impressive capacity to process and generate text across various domains, it encounters challenges when confronted with extensive or high-precision tasks. One noteworthy observation was the model's inclination to repeat certain text segments when subjected to complex or extended generation requests. While this repetition may seem trivial in broader contexts, it can pose issues within the specific scope of YAML file generation for Ansible.

Therefore, it is crucial to recognize that, while LLaMA 7B stands as a technological achievement, its application in precision-demanding domains, such as the creation of Ansible files, necessitates thorough evaluation. In a professional environment where every detail matter, it is our responsibility to ensure that the tools we employ are ideally suited to the tasks they are meant to fulfill.

\subsection{Conclusion}
Face à notre problème de génération de fichiers Ansible en YAML avec une certaine précision et fiabilité, il est impératif de reconsidérer l'outil. Bien que LLaMA 7B soit une prouesse technologique en soi, son application dans le domaine spécifique de la génération de fichiers Ansible est plus compliquer. La complexité et la finesse requises pour cette tâche pourraient bénéficier d'un modèle doté d'une capacité de traitement supérieure et d'une gestion du contexte plus sophistiquée. Il serait judicieux d'explorer l'utilisation d'un modèle de langage plus avancé et robuste que LLaMA 7B. Un tel modèle, armé d'une meilleure compréhension contextuelle, serait en mesure de produire des fichiers YAML meilleurs sur le plan syntaxique, mais également sur l'intégration avec Ansible.

Faced with our challenge of generating YAML Ansible files with a certain level of precision and reliability, it is imperative to reconsider the tool. While LLaMA 7B is a technological success, its application in the specific domain of Ansible file generation is more contestable. The complexity and precision required for this task could benefit from a model with superior processing capacity and more sophisticated context management. It would be prudent to explore the utilization of a more advanced and robust language model than LLaMA 7B. Such a model, armed with a better contextual understanding, would be capable of producing YAML files that excel not only in syntax but also in integration with Ansible.

In conclusion, while we acknowledge the capabilities of LLaMA 7B, the imperative for production reliability urges us to consider more suitable alternatives to meet the requirements of generating Ansible YAML files.

\vspace{0.5cm}
\section{LLaMA 13b}
\subsection{Introduction to Model}
Within the domain of Large-Scale Language Models (LLMs), LLaMA 13B, designed by Meta AI~\cite{touvron2023llama}, emerges as a significant advancement compared to its counterpart, LLaMA 7B. LLaMA 13B is distinguished by particular features that render it more suitable for specialized professional applications.

With its 13 billion parameters, LLaMA 13B aims to outperform previous models in terms of performance while optimizing operational efficiency. According to the studied scientific documentation, LLaMA 13B excels in numerous benchmarks compared to GPT-3 (175B), despite its significantly reduced size. This achievement is all the more remarkable as, like LLaMA 7B, LLaMA 13B was trained exclusively on open datasets, ensuring greater transparency.

The primary strength of LLaMA 13B resides in its ability to yield first-rate outcomes while being ten times smaller than some of its competitors. This compactness, complemented by its high performance, positions it as a plausible solution for businesses where responsiveness and precision hold significance. Furthermore, its capacity to function on a single GPU eases its incorporation into various scenarios, thereby enhancing the accessibility of top-tier LLMs.

\subsection{Observations}
When comparing LLaMA-7B to LLaMA-13B, the primary difference lies in the number of parameters. However, this quantitative increase results in a significantly improved quality of LLaMA-13B's capabilities. Indeed, this latter model demonstrates a superior ability to generate text with increased precision. Nonetheless, it is essential to underscore that, although the YAML generated by LLaMA-13B maintains the structural appearance of YAML, its content occasionally diverges from the expected semantics.

\subsection{Lesson}
LLaMA-13B stands out by revealing the substantial influence of parameter count on model quality due to its improved performance when compared to its counterpart, LLaMA-7B. Its refined architecture illustrates that the augmentation of parameters can result in remarkable strides in precision. Nonetheless, despite its diminutive size, LLaMA-13B remains capable of competing with behemoths such as GPT-3 (175B). Consequently, it seems to offer a practical solution for efficiency and performance focused enterprises.

\subsection{Conclusion}
While LLaMA-13B represents a significant advancement over its counterpart, LLaMA-7B, in terms of capabilities and precision, it is essential to note that YAML file generation remains a challenge. Despite the impressive progress made by LLaMA-13B, our experimentation shows that the generated YAML files still do not fully meet our quality requirements.

\vspace{0.5cm}
\section{Alpaca 13B}
\subsection{Introduction to Model}
Alpaca 13b is a large-scale language model developed by the Stanford Center for Research in Foundation Models (CRFM)\cite{Alpaca}\cite{GitAlpaca}. It represents an evolution of the Llama 13b model, benefiting from additional training to enhance its performance. Specifically, Alpaca was trained by collecting responses from the text-davinci-3.5 model to a set of queries. These data were subsequently used for fine-tuning Llama, resulting in the creation of Alpaca.

The Alpaca 13b model has been purpose-built to comprehend and respond with greater precision to queries, owing to the extra training it has received. This capability makes it exceptionally suitable for specific tasks, preventing the model from veering into overly general and less relevant responses.

\subsection{Observations}
Through our in-depth experiments, we have observed that the Alpaca 13b model exhibits distinct performance, especially in generating precise responses, compared to Llama 13b. This difference is noticeable, particularly when Alpaca 13b is faced with queries. This capability can be attributed to the structure of the queries used during the fine-tuning of Alpaca. This specific training seems to guide Alpaca 13b toward better alignment with queries, reducing the likelihood of errors such as responses repeating the input or being completely off-topic. However, the model still encounters challenges in our task of generating YAML files for Ansible.

A signifiant example of YAML code generation carried out by Alpaca 13B can be found in the appendix.

\subsection{Lesson}
The experience with Alpaca 13b highlitght the importance of fine-tuning in language model development. Even when fine-tuning is not carried out directly on target data, it can greatly enhance the model's response quality. This implies that adapting a model to specific tasks, even indirectly, can result in substantially improved performance.

\subsection{Conclusion}
Fine-tuning is an essential component to ensure the relevance of a language model's responses. Hence, it is imperative to carefully choose the data on which the model is refined. With Alpaca 13b, we possess a potent tool that, owing to its specialized training, can produce more precise and pertinent responses. Nevertheless, despite these enhancements, the Alpaca 13b model still cannot generate YAML files for Ansible with adequate accuracy.

\vspace{0.5cm}
\section{LLaMA-2 13B}
\subsection{Introduction to Model}
LLaMA-2 \cite{Gitllma} \cite{touvron2023llama2}, the most recent iteration of the LLaMA model, represents a major accomplishment by Meta, previously known as Facebook. LLaMA-2 differentiates itself from its earlier version through a series of substantial improvements. It not only offers heightened relevance in its responses but also exhibits an extended contextual capacity, capable of handling up to 4K tokens. This reinforced capability is especially advantageous for applications requiring in-depth comprehension, such as the generation of YAML files for Ansible.

A critical aspect of LLaMA-2 is its chat-based training. Models fine-tuned for chat are specifically designed for dialogue applications. To achieve the expected characteristics and performance of these models, a specific formatting, as defined in the chat-completion documentation, must be adhered to. However, what makes LLaMA-2 particularly unique is its training approach. The model has benefited from human feedback, refining its responses and enhancing its relevance. This human interaction has played a pivotal role in model optimization, enabling it to better comprehend and address complex queries while maintaining a human nuance in its responses.

In summary, LLaMA-2 is more than just an update to its predecessor; it represents a comprehensive redesign, integrating technological advancements and human feedback to deliver a state-of-the-art tool in the field of artificial intelligence.

\subsection{Observations}
In the realm of our technological endeavors, we have opted to integrate the chat version of LLaMA-2. This specific variant of the model has undergone precise tuning, designed to emulate and engage in interactions. This meticulous adjustment has yielded remarkably relevant responses, showcasing a deep comprehension of the presented queries.

During our tests, we observed that the YAML code generated by the model began to exhibit a more coherent structure and logic, nearing the expected standard. While some segments of the generated code still require adjustments to be fully functional, the improvement is undeniable and promising.

Furthermore, the extension of LLaMA-2's context capacity to 4K tokens has opened new avenues for prompt engineering. This expanded capability allowed us to integrate YAML code snippets directly into the prompt, albeit succinctly. This strategy has proven to be advantageous, as it has positively influenced the quality of outputs, making the results more in line with our professional expectations.

As an appendix to this publication, we have included three examples of YAML generation produced by LLaMA-2.

The observation reveals that, in all the provided examples, there is an undesired repetition of the prompt in relation to "Remove unsecure packages." Additionally, the first provided example exhibits multiple code sections, making user selection more complex. Nevertheless, it's important to note that this repetition phenomenon occurred only once in the second provided example. It's also evident that this phenomenon didn't manifest in the third example. Consequently, the results are deemed acceptable from a human perspective, but the non-idempotent output character would necessitate substantial additional development efforts for industrial deployment.

\subsection{Lesson}
It will not surprise the reader, but we believe that the constant evolution of language models, such as that observed with LLaMA-2, reminds us of the importance of ongoing research and innovation. In this regard, we would gladly welcome new research and innovation credits ;-). Returning to our subject, when we examine LLaMA-2's intrinsic capabilities, it is undeniable that its performance is remarkable, especially after fine-tuning specifically geared towards chat-type interactions or context enlargement. These optimizations bestow upon it an ability to respond with increased precision and relevance.

However, as we approach the complex challenge of YAML file generation, we encounter a nuanced reality. While LLaMA-2 excels in many domains, precise YAML generation reminds us that each task has its specificities and subtleties. This, by no means, diminishes LLaMA-2's achievements but rather underscores the significance of adaptability and specialization in tuning and model. The model does, in fact, consistently generate prompt repetition or grammar errors, particularly in the challenging realm of Ansible modules.

\subsection{Conclusion}
LLaMA-2 is an exceptional model that has demonstrated its potential in our numerous tests. However, in our context of YAML Ansible file generation, while the results are promising, they are not yet entirely satisfactory.

\vspace{0.5cm}
\section{CodeUp 13B chat}
\subsection{Introduction to Model}
CodeUp \cite{codeup} is an open-source model purposefully crafted for the coding domain, with a primary emphasis on high-quality instruction data for code generation.

CodeUp is an evolution of the Llama2 13b model, with a general improvement approach quite similar to the Alpaca model\cite{Alpaca}. The LLaMA-13B model has undergone fine-tuning to specialize in the field of code. This specialization was achieved using an RTX 3090 graphics card, thereby demonstrating the feasibility of training models of this magnitude on consumer-grade hardware.

A fresh round of training using enhanced input data is currently underway (\url{https://huggingface.co/datasets/rombodawg/Legacy_MegaCodeTraining200k}).

\subsection{Observations}
The distinction between Llama2 13b and CodeUp 13b is particularly striking. Delving into the intricacies of code generation, we observe that CodeUp 13b excels, even in languages like YAML, which were not part of its training. This performance is all the more remarkable when considering the hardware resources on which the model was fine-tuned. The use of such accessible hardware as the RTX 3090, yet capable of producing such sophisticated results, attests to the optimized efficiency of CodeUp and the soundness of LlaMA-2 team's choices.

Furthermore, this observation underscores the importance of specialization in the field of code. While many models concentrate on a broader range of applications, CodeUp's targeted focus enables it to fill the gaps left by more generalist models. For businesses seeking to integrate artificial intelligence solutions into their processes, this specialization provides invaluable added value, ensuring increased accuracy and relevance in code-related tasks.

\subsection{Lesson}
The efficacy of fine-tuning, when executed with precision and expertise, becomes apparent in the outcomes achieved. CodeUP is a recent exemplar of this. This technique, though it may appear deceptively simple, possesses the potential to profoundly enhance a model's performance, making it more apt and pertinent for specific tasks. In the professional context, where precision and relevance are of paramount importance, the significance of such optimizations cannot be underestimated. By investing in focused fine-tuning methods, companies can not only enhance the quality of their AI-based tools but also realize significant efficiency gains. Undoubtedly, this is the path to pursue to resolve our issue.

\subsection{Conclusion}
CodeUp stands out for its ability to produce YAML of nearly optimal quality. However, it is worth noting that, despite its impressive performance, the model exhibits certain shortcomings, particularly regarding the utilization of specific modules. Hence, a training regimen with greater emphasis on the grammar and spelling of Ansible modules is likely needed to ensure the results reach an adequate level of quality for production considerations.

\vspace{0.8cm}
\section{Study findings}
From our experience with CodeUp and all our testing, the unequivocal significance of meticulously tailored fine-tuning has emerged. When this refinement is thoughtfully coupled with advancements in context and prompt design, the achieved outcomes far exceed those of prior models like Llama 7b. This finding not only underscores the potential of approaches such as Alpaca and, in particular, CodeUp for enterprises aiming to automate code generation but also the continued and auspicious evolution of large-scale language models in the professional sphere.

\vspace{0.5cm}
\section{What's next ?}
\subsection{The quest to optimization}
Transformer architectures have established their dominance. However, along with their success comes the challenge of managing the increasing complexity of these models, which can result in demands for computational power and memory. Moreover, it is apparent that there is now an imperative need for drastic optimization to align with embedded and personal computing requirements, as well as energy considerations. This is where innovative optimization approaches, such as low-rank matrix techniques~\cite{huang2021shuffle} and Linformer models~cite{wang2020linformer}, come into play to enhance the efficiency and scalability of Transformers. Low-rank matrix techniques focus on reducing complexity by identifying redundant structures or linear relationships among model weight matrix elements. By utilizing low-rank matrix approximations, these approaches reduce the number of parameters and necessary operations, leading to a significant reduction in computational and memory resources.

The Linformer model addresses the efficiency challenge of Transformers by reducing unnecessary self-attention computations across distant positions. Instead of calculating attention scores between every pair of positions, Linformer focuses on local relationships, employing a linear approximation to estimate distant connections. This approach significantly diminishes the computational cost of attention while maintaining an acceptable prediction quality.

These optimized approaches, whether based on low-rank matrices or attention approximations, offer significant advantages in terms of speed and efficiency for Transformers. They accelerate training and inference, expand the scope of potential applications, and facilitate deployment on resource-constrained environments. While striking a balance between complexity and performance, these optimized approaches align with the current trend in research and development aimed at making machine learning more accessible and feasible in the real world.

\subsection{Explainability}
Explainability in Artificial Intelligence (AI) has emerged as a pivotal pursuit in the development of intricate and sophisticated models, including deep neural networks and transformer architectures. While these models demonstrate remarkable performance across a broad spectrum of tasks, their inner workings often remain enigmatic, prompting fundamental inquiries regarding trust, accountability, and comprehension of their decisions.

Explainability aims to address these concerns by providing means to make AI models more transparent and comprehensible for practitioners, researchers, and users. One approach involves unveiling how models arrive at their decisions by scrutinizing the relationships between inputs and outputs, as well as the internal mechanisms leading to these outcomes. This serves to demystify the decision-making process of complex models and identify the factors influencing their predictions.

An illustrative example of the significance of explainability in elucidating the operations of large-scale Transformer architectures lies in the Neural Tangent Kernel (NTK) \cite{jacot2020neural}. The NTK serves as a theoretical instrument offering a distinctive perspective on the transformations of deep neural networks, including Transformers, as they scale substantially. In the context of Transformers, the NTK contributes to the dissection and analysis of the relationships between inputs, model parameters, and outputs, especially as the model's size approaches infinity.

Utilizing the NTK, researchers have achieved a comprehensive understanding of how pivotal mechanisms like attention and feed-forward work in concert to capture intricate relationships within data. The NTK reveals asymptotic trends, providing insights into the model's behavior as parameters increase significantly. This theoretical approach illuminates the characteristics and boundaries of Transformers, thus contributing to a deeper comprehension of their functioning at scale.

To conclude, explainability in AI, as exemplified by theoretical tools like the Neural Tangent Kernel, plays a vital role in enhancing the accessibility and comprehensibility of cutting-edge architectures such as Transformers. It provides a means to interpret intricate interactions among inputs, parameters, and outputs, while also offering insights into how these models perform at a large scale. This pursuit of explainability is crucial for establishing trust in AI and for advocating ethical and responsible use of these potent technological tools.

\end{multicols}

\vspace{5cm}
\bibliography{biblio}{}
\bibliographystyle{plain}

\newpage
\section{Annexes}
\subsection{Annexe 1: Llama 7b}

\begin{lstlisting}[language=IAChat, escapeinside={(*@}{@*)}]
[PROMPT] :
Below is an instruction that describes a task, paired with an input that provides further context. Write a response that appropriately completes the request.

### Instruction:
Make a details plan of how to configure an architecture of a ansible yaml file base on the input as an expert in Ansible with Yaml language. It should explain in detaile how the file should be structured describing each tasks with a small description that should include the name of the task, what it should do and what function it should use.
### Input:
Ansible Yaml file should contain tasks that are intended to be used for the role 'role_anssi_linux'. The tasks should include remount_all, reload_sysctl, restart_auditd, restart_fail2ban, restart_sshd, restart_firewalld, and if possible, using Ansible built-in functions. The arguments for these functions should be specified, if applicable.

### Response:
[MODEL] : The response should include a YAML file which should include tasks that are intended to be used for the role 'role_anssi_linux'. The tasks should include remount_all, reload_sysctl, restart_auditd, restart_fail2ban, restart_sshd, and if possible, using Ansible built-in functions. The arguments for these functions should be specified, if applicable.
\end{lstlisting}

\subsection{Annexe 2: Llama 13B}
\subsubsection{Tir 1}

\begin{lstlisting}[language=IAChat, escapeinside={(*@}{@*)}]
[PROMPT] :
Below is an instruction that describes a task, paired with an input that provides further context. Write a response that appropriately completes the request.

### Instruction:
Make a details plan of how to configure an architecture of a ansible yaml file base on the input as an expert in Ansible with Yaml language. It should explain in detaile how the file should be structured describing each tasks with a small description that should include the name of the task, what it should do and what function it should use.

### Input:
Ansible Yaml file should contain tasks that are intended to be used for the role \'role_anssi_linux\'. The tasks should include remount_all, reload_sysctl, restart_auditd, restart_fail2ban, restart_sshd, restart_firewalld, and if possible, using Ansible built-in functions. The arguments for these functions should be specified, if applicable.

### Response:
[MODEL] : You can use the following structure for your response:

    {
    "response": {
     "status": "success",
     "plan": "The following is an example of what you need to do. You have to provide a complete ansible file, that should contain at least the following tasks:
     <task1>
     <task2>
     <task3>
     <task4>

### Explanation:
Ansible is a tool for automating systems administration, network configuration and other manual tasks.

- **status**: Describe your task.
- **plan**: Provide a detailed description of your task.

### Challenge:
1. Create a file that should contain at least the following tasks.

### Submissions:
- Response should be in markdown form.

\end{lstlisting}

\subsection{Annexe 3: Alpaca 13b}
\subsubsection{Tir 1}
\begin{lstlisting}[language=IAChat, escapeinside={(*@}{@*)}]
[PROMPT] :
Below is an instruction that describes a task, paired with an input that provides further context. Write a response that appropriately completes the request.

### Instruction:
Make a details plan of how to configure an architecture of a ansible yaml file base on the input as an expert in Ansible with Yaml language. It should explain in detaile how the file should be structured describing each tasks with a small description that should include the name of the task, what it should do and what function it should use.

### Input:
Ansible Yaml file should contain tasks that are intended to be used for the role 'role_anssi_linux'. The tasks should include remount_all, reload_sysctl, restart_auditd, restart_fail2ban, restart_sshd, restart_firewalld, and if possible, using Ansible built-in functions. The arguments for these functions should be specified, if applicable.

### Response:

[MODEL] :
An Ansible Yaml file for the role 'role_anssi_linux' should contain the following tasks: 

- remount_all: This task should be used to remount all of the file systems in the /etc/fstab file. It should take the /etc/fstab as an argument. 

- reload_sysctl: This task should be used to reload the /etc/sysctl.conf file. It should take the /etc/sysctl.conf as an argument. 

- restart_auditd: This task should be used to restart the auditd service. It should take the auditd service as an argument.

- restart_fail2ban: This task should be used to restart the fail2ban service. It should take the fail2ban service as an argument.

- restart_sshd: This task should be used to restart the sshd service. It should take the sshd service as an argument.

- restart_firewalld: This task should be used to restart the firewalld service. It should take the firewalld service as an argument. 

- if_possible: This task should be used to execute a certain task if it is possible. It should take the task as an argument.

\end{lstlisting}

\subsubsection{Tir 2}
\begin{lstlisting}[language=IAChat, escapeinside={(*@}{@*)}]
[PROMPT] :
Below is an instruction that describes a task, paired with an input that provides further context. Write a response that appropriately completes the request.

### Instruction:
Write a yml file in Yaml language in Ansible syntaxe, that folow the plan of configuration given in input. The responce should only be the content of a yml file.

### Input:
An Ansible Yaml file for the role 'role_anssi_linux' should contain the following tasks: 

- remount_all: This task should be used to remount all of the file systems in the /etc/fstab file. It should take the /etc/fstab as an argument. 

- reload_sysctl: This task should be used to reload the /etc/sysctl.conf file. It should take the /etc/sysctl.conf as an argument. 

- restart_auditd: This task should be used to restart the auditd service. It should take the auditd service as an argument.

- restart_fail2ban: This task should be used to restart the fail2ban service. It should take the fail2ban service as an argument.

- restart_sshd: This task should be used to restart the sshd service. It should take the sshd service as an argument.

- restart_firewalld: This task should be used to restart the firewalld service. It should take the firewalld service as an argument. 

- if_possible: This task should be used to execute a certain task if it is possible. It should take the task as an argument.

### Response:

[MODEL] :
role_anssi_linux:
  remount_all:
    - args: /etc/fstab
  reload_sysctl:
    - args: /etc/sysctl.conf
  restart_auditd:
    - args: auditd
  restart_fail2ban:
    - args: fail2ban
  restart_sshd:
    - args: sshd
  restart_firewalld:
    - args: firewalld

\end{lstlisting}

\subsection{Annexe 4: Llama 2 13b}
\subsubsection{Tir 1}
\begin{lstlisting}[language=IAChat, escapeinside={(*@}{@*)}]
[PROMPT] :
    [INST] <<SYS>>
    You are a helpful, respectful and honest assistant. Always answer as helpfully as possible, while being safe.  Your answers should not include any harmful, unethical, racist, sexist, toxic, dangerous, or illegal content. Please ensure that your responses are socially unbiased and positive in nature.
    You also are an expert in Yaml language and Ansible.
    <</SYS>>

    Make a details plan of how to configure an architecture of a ansible yaml file. It should explain in detaile how the file should be structured describing each tasks with a small description.
    The Ansible Yaml file should contain tasks that are intended to be used for the role \'role_anssi_linux\'. The tasks should include remount_all, reload_sysctl, restart_auditd, restart_fail2ban, restart_sshd, restart_firewalld, and if possible, using Ansible built-in functions. The arguments for these functions should be specified, if applicable.
    Here is an example of Ansible task in Yaml but don\'t include it if you don\'t need it:
    ```yaml
    ---
    - name: unistall unsecure packages
      ansible.builtin.package:
        name: "{{ unsecure_packages_list }}"
        state: absent
    ```
    Then write this Yaml file.  [/INST]

[MODEL] :
    Sure, I can help you with that! Here\'s a detailed plan for configuring an Ansible YAML file to perform various tasks for a Linux-based system. The tasks included in this plan are intended for the role \'role_anssi_linux\'.
    **Structure of the YAML file**
    The YAML file should have a root element named \'ansible\', which is the default namespace for Ansible. Within the \'ansible\' element, we will define a list of tasks that need to be performed. Each task will be defined within a \'task\' element, and will include a \'name\' attribute to specify the task name, and a \'state\' attribute to specify the desired state of the task (e.g., \'present\' or \'absent\').
    Here\'s an example of the structure of the YAML file:
    ```yaml
    ---
    - name: Configure unsecure packages
      ansible.builtin.package:
        name: "{{ unsecure_packages_list }}"
        state: absent

    - name: Remount all file systems
      ansible.builtin.mount:
        state: remount

    - name: Reload sysctl
      ansible.builtin.sysctl:
        state: reload

    - name: Restart auditd
      ansible.builtin.service:
        name: auditd
        state: restart

    - name: Restart fail2ban
      ansible.builtin.service:
        name: fail2ban
        state: restart

    - name: Restart sshd
      ansible.builtin.service:
        name: sshd
        state: restart

    - name: Restart firewalld
      ansible.builtin.service:
        name: firewalld
        state: restart
    ```
    **Task 1: Uninstall unsecure packages**

    This task will remove any unsecure packages that are currently installed on the system. The \'ansible.builtin.package\' module will be used to specify the packages to be removed, and the \'state\' attribute will be set to \'absent\' to ensure that the packages are completely removed.
    ```yaml
    - name: unistall unsecure packages
      ansible.builtin.package:
        name: "{{ unsecure_packages_list }}"
        state: absent
    ```
    **Task 2: Remount all file systems**

    This task will remount all file systems to ensure that any changes made to the system take effect immediately. The \'ansible.builtin.mount\' module will be used to specify the file systems to be remounted, and the \'state\' attribute will be set to \'remount\' to ensure that the file systems are remounted.
    ```yaml
    - name: Remount all file systems
      ansible.builtin.mount:
        state: remount
    ```
    **Task 3: Reload sysctl**

    This task will reload the sysctl configuration to ensure that any changes made to the system take effect immediately. The \'ansible.builtin.sysctl\' module will be used to specify the sysctl configuration to be reloaded, and the \'state\' attribute will be set to \'reload\' to ensure that the sysctl configuration is reloaded.
    ```yaml
    - name: Reload sysctl
      ansible.builtin.sysctl:
        state: reload
    ```
    **Task 4: Restart auditd**

    This task will restart the auditd service to ensure that any changes made to the system take effect immediately. The \'ansible.builtin.service\' module will be used to specify the service to be restarted, and the \'state\' attribute will be set to \'restart\' to ensure that the service is restarted.
    ```yaml
    - name: Restart auditd
      ansible.builtin.service:
        name: auditd
        state: restart
    ```
    **Task 5: Restart fail2ban**

    This task will restart the fail2ban service to ensure that any changes made to the system take effect immediately. The \'ansible.builtin.service\' module will be used to specify the service to be restarted, and the \'state\' attribute will be set to \'restart\' to ensure that the service is restarted.
    ```yaml
    - name: Restart fail2ban
      ansible.builtin.service:
        name: fail2ban
        state: restart
    ```
    **Task 6: Restart sshd**

    This task will restart the sshd service to ensure that any changes made to the system take effect immediately. The \'ansible.builtin.service\' module will be used to specify the service to be restarted, and the \'state\' attribute will be set to \'restart\' to ensure that the service is restarted.
    ```yaml
    - name: Restart sshd
      ansible.builtin.service:
        name: sshd
        state: restart
    ```
    **Task 7: Restart firewalld**

    This task will restart the firewalld service to ensure that any changes made to the system take effect immediately. The \'ansible.builtin.service\' module will be used to specify the service to be restarted, and the \'state\' attribute will be set to \'restart\' to ensure that the service is restarted.
    ```yaml
    - name: Restart firewalld
      ansible.builtin.service:
        name: firewalld
        state: restart
    ```
    **Additional tasks (optional)**

    In addition to the tasks listed above, you may also want to include additional tasks to perform other actions, such as updating the system\'s package list, checking for updates, or disabling unnecessary services.
    Here\'s an example of an additional task to update the system\'s package list:
    ```yaml
    - name: Update package list
      ansible.builtin.package:
        name: update
        state: present
    ```
    You can include this task within the same YAML file as the other tasks, or you can create a separate YAML file for this task.
    **Conclusion**

    In conclusion, this YAML file will contain a list of tasks that will be performed on a Linux-based system to configure the system and ensure that it is secure and up-to-date. The tasks included in this plan are intended for the role \'role_anssi_linux\', and they will be performed using the Ansible built-in modules. The YAML file will be structured in a way that is easy to read and understand, with each task described in detail and including any necessary arguments or parameters.

\end{lstlisting}

\subsubsection{Tir 2}
\begin{lstlisting}[language=IAChat, escapeinside={(*@}{@*)}]
[PROMPT] :
    [INST] <<SYS>>
    You are a helpful, respectful and honest assistant. Always answer as helpfully as possible, while being safe.  Your answers should not include any harmful, unethical, racist, sexist, toxic, dangerous, or illegal content. Please ensure that your responses are socially unbiased and positive in nature.
    You also are an expert in Yaml language and Ansible.
    <</SYS>>

    Make a details plan of how to configure an architecture of a ansible yaml file. It should explain in detaile how the file should be structured describing each tasks with a small description.
    The Ansible Yaml file should contain tasks that are intended to be used for the role \'role_anssi_linux\'. The tasks should include remount_all, reload_sysctl, restart_auditd, restart_fail2ban, restart_sshd, restart_firewalld, and if possible, using Ansible built-in functions. The arguments for these functions should be specified, if applicable.
    Here is an example of Ansible task in Yaml but don\'t include it if you don\'t need it:
    ```yaml
    ---
    - name: unistall unsecure packages
      ansible.builtin.package:
        name: "{{ unsecure_packages_list }}"
        state: absent
    ```
    Then write this Yaml file.  [/INST]

[MODEL] :
   Certainly! Here is a detailed plan for configuring an Ansible YAML file to perform various tasks on a Linux system. The file should be structured as follows:
    ```yaml
    ---
    - name: Role: anssi_linux
      hosts: all
      become: true

      tasks:
        - name: Remount all file systems
          ansible.builtin.command:
            cmd: "remount all"

        - name: Reload sysctl
          ansible.builtin.command:
            cmd: "sysctl -r"

        - name: Restart auditd
          ansible.builtin.service:
            name: auditd
            state: restarted

        - name: Restart fail2ban
          ansible.builtin.service:
            name: fail2ban
            state: restarted

        - name: Restart sshd
          ansible.builtin.service:
            name: sshd
            state: restarted

        - name: Restart firewalld
          ansible.builtin.service:
            name: firewalld
            state: restarted

        - name: Install unsecure packages
          ansible.builtin.package:
            name: "{{ unsecure_packages_list }}"
            state: present
    ```
    Here is a brief description of each task:

    1. Remount all file systems: This task remounts all file systems to ensure that they are properly mounted and accessible.
    2. Reload sysctl: This task reloads the sysctl configuration to ensure that any changes made to the configuration are applied.
    3. Restart auditd: This task restarts the auditd service to ensure that it is functioning properly.
    4. Restart fail2ban: This task restarts the fail2ban service to ensure that it is functioning properly.
    5. Restart sshd: This task restarts the sshd service to ensure that it is functioning properly.
    6. Restart firewalld: This task restarts the firewalld service to ensure that it is functioning properly.
    7. Install unsecure packages: This task installs a list of unsecure packages that are known to pose a security risk. The list of packages should be specified in the YAML file using the `unsecure_packages_list` variable.
    The `ansible.builtin.command` and `ansible.builtin.service` modules are used to execute commands and manage services, respectively. The `become` keyword is used to elevate the privileges of the Ansible task to perform the tasks as the root user.
    This plan should provide a comprehensive set of tasks to configure and secure a Linux system using Ansible. Please note that the specific tasks and arguments may vary depending on the specific requirements of your system.
\end{lstlisting}

\subsubsection{Tir 3}
\begin{lstlisting}[language=IAChat, escapeinside={(*@}{@*)}]
[PROMPT] :
    [INST] <<SYS>>
    You are a helpful, respectful and honest assistant. Always answer as helpfully as possible, while being safe.  Your answers should not include any harmful, unethical, racist, sexist, toxic, dangerous, or illegal content. Please ensure that your responses are socially unbiased and positive in nature.
    You also are an expert in Yaml language and Ansible.
    <</SYS>>

    Make a details plan of how to configure an architecture of a ansible yaml file. It should explain in detaile how the file should be structured describing each tasks with a small description.
    The Ansible Yaml file should contain tasks that are intended to be used for the role \'role_anssi_linux\'. The tasks should include remount_all, reload_sysctl, restart_auditd, restart_fail2ban, restart_sshd, restart_firewalld, and if possible, using Ansible built-in functions. The arguments for these functions should be specified, if applicable.
    Here is an example of Ansible task in Yaml but don\'t include it if you don\'t need it:
    ```yaml
    ---
    - name: unistall unsecure packages
      ansible.builtin.package:
        name: "{{ unsecure_packages_list }}"
        state: absent
    ```
    Then write this Yaml file.  [/INST]

[MODEL] :
    Sure, I can help you with that! Here\'s a detailed plan for configuring an Ansible YAML file to perform various tasks for a Linux-based system. The file will be structured as follows:
    ---
    - name: Role for Linux
      hosts: all
      become: true

    # Remount all file systems read-only
      tasks:
        - name: Remount all file systems read-only
          ansible.builtin.mount:
            mountpoint: "/"
            state: remount
            read_only: true

    # Reload sysctl
      tasks:
        - name: Reload sysctl
          ansible.builtin.sysctl:
            name: "net.ipv4.ip_non_ext_hash"
            value: "0"
            state: reload

    # Restart services
      tasks:
        - name: Restart auditd
          ansible.builtin.service:
            name: auditd
            state: restart

        - name: Restart fail2ban
          ansible.builtin.service:
            name: fail2ban
            state: restart

        - name: Restart sshd
          ansible.builtin.service:
            name: sshd
            state: restart

        - name: Restart firewalld
          ansible.builtin.service:
            name: firewalld
            state: restart

    # Remove unsecure packages
      tasks:
        - name: Remove unsecure packages
          ansible.builtin.package:
            name: "{{ unsecure_packages_list }}"
            state: absent

    Note that this is just an example file, and you should modify it to suit your specific needs. Here\'s a brief description of each task:
    1. Remount all file systems read-only: This task remounts all file systems read-only to prevent changes to the file system during the deployment.
    2. Reload sysctl: This task reloads the sysctl configuration to ensure that the changes take effect.
    3. Restart services: These tasks restart the auditd, fail2ban, sshd, and firewalld services to ensure that they are running with the updated configuration.
    4. Remove unsecure packages: This task removes any unsecure packages that are listed in the unsecure_packages_list variable.
    The arguments for each task are specified as follows:
    * ansible.builtin.mount:
    \t+ mountpoint: "/" (the root mount point)
    \t+ state: remount (to remount all file systems read-only)
    \t+ read_only: true (to mount the file systems read-only)
    * ansible.builtin.sysctl:
    \t+ name: "net.ipv4.ip_non_ext_hash" (the sysctl configuration to reload)
    \t+ value: "0" (the new value for the sysctl configuration)
    \t+ state: reload (to reload the sysctl configuration)
    * ansible.builtin.service:
    \t+ name: (the name of the service to restart)
    \t+ state: restart (to restart the service)

    In the example file, the unsecure_packages_list is not specified, so no packages will be removed. You can modify this variable to list the packages that you want to remove.
    That\'s it! This is a basic plan for configuring an Ansible YAML file to perform various tasks for a Linux-based system. You can modify the file to suit your specific needs and add more tasks as needed.

\end{lstlisting}

\subsection{Annexe 5: CodeUp 13b}
\subsubsection{Tir 1}
\begin{lstlisting}[language=IAChat, escapeinside={(*@}{@*)}]
[PROMPT] :
Below is an instruction that describes a task. Write a response that appropriately completes the request.

### Instruction:
 Make a details plan of how to configure an architecture of a ansible yaml file. It should explain in detaile how the file should be structured describing each tasks with a small description.
The Ansible Yaml file should contain tasks that are intended to be used for the role \'role_anssi_linux\'. The tasks should include remount_all, reload_sysctl, restart_auditd, restart_fail2ban, restart_sshd, restart_firewalld, and if possible, using Ansible built-in functions. The arguments for these functions should be specified, if applicable.
Here is an example of Ansible task in Yaml but don\'t include it if you don\'t need it:
```yaml
---
- name: unistall unsecure packages
  ansible.builtin.package:
    name: "{{ unsecure_packages_list }}"
    state: absent
```
Then write this Yaml file. 

### Response:
[MODEL] :
Here is a detailed plan for configuring the architecture of an Ansible YAML file to install and configure a Linux server with the role \'role_anssi_linux\':

1. Overview

The following YAML file will install and configure a Linux server with the role \'role_anssi_linux\'. The file will include tasks for remounting all file systems, reloading sysctl, restarting auditd, fail2ban, sshd, firewalld, and other relevant services.

2. Tasks

### Remount all file systems

The first task will remount all file systems read-only to ensure that no changes can be made to the file system while the role is being applied. This is important to prevent accidental changes or modifications to the file system, which could compromise the security of the system.
```yaml
- name: Remount all file systems read-only
  ansible.builtin.file:
    path: "/"
    state: read-only
```

### Reload sysctl

The next task will reload the sysctl configuration to ensure that any changes made to the configuration are immediately applied.
```yaml
- name: Reload sysctl
  ansible.builtin.sysctl:
    name: "{{ sysctl_name }}"
    state: restart
```

### Restart auditd

The next task will restart the auditd service to ensure that any changes made to the configuration are immediately applied.
```yaml
- name: Restart auditd
  ansible.builtin.service:
    name: auditd
    state: restart
```

### Restart fail2ban

The next task will restart the fail2ban service to ensure that any changes made to the configuration are immediately applied.
```yaml
- name: Restart fail2ban
  ansible.builtin.service:
    name: fail2ban
    state: restart
```

### Restart sshd

The next task will restart the sshd service to ensure that any changes made to the configuration are immediately applied.
```yaml
- name: Restart sshd
  ansible.builtin.service:
    name: sshd
    state: restart
```

### Restart firewalld

The next task will restart the firewalld service to ensure that any changes made to the configuration are immediately applied.
```yaml
- name: Restart firewalld
  ansible.builtin.service:
    name: firewalld
    state: restart
```

### Install unsecure packages

The next task will uninstall any unsecure packages that are installed on the system. This is important to prevent potential security vulnerabilities.
```yaml
- name: Uninstall unsecure packages
  ansible.builtin.package:
    name: "{{ unsecure_packages_list }}"
    state: absent
```

3. Conclusion

The above YAML file will install and configure a Linux server with the role \'role_anssi_linux\'. The file includes tasks for remounting all file systems, reloading sysctl, restarting auditd, fail2ban, sshd, firewalld, and other relevant services. The file also includes a task for uninstalling any unsecure packages that are installed on the system.

I hope this helps! Let me know if you have any questions or if you need further assistance.

\end{lstlisting}

\end{document}